\documentclass[10pt,twocolumn,letterpaper]{article}

\usepackage[pagebackref=true,breaklinks=true,letterpaper=true,colorlinks,bookmarks=false]{hyperref}

\usepackage{cvpr}
\usepackage{times}
\usepackage{epsfig}
\usepackage{graphicx}
\usepackage{amsmath}
\usepackage{amssymb}

\usepackage{listings}

\usepackage{caption}

\cvprfinalcopy 

\def\cvprPaperID{****} 

\begin{document}

\title{Seeing What Is Not There:\\ Learning Context to Determine Where Objects Are Missing}

\author{Jin Sun \qquad David W. Jacobs\\
Department of Computer Science\\
University of Maryland\\
{\tt\small \{jinsun,djacobs\}@cs.umd.edu}
}

\twocolumn[{%
\renewcommand\twocolumn[1][]{#1}%
\maketitle
\begin{center}
    \centering
    \includegraphics[width=\textwidth]{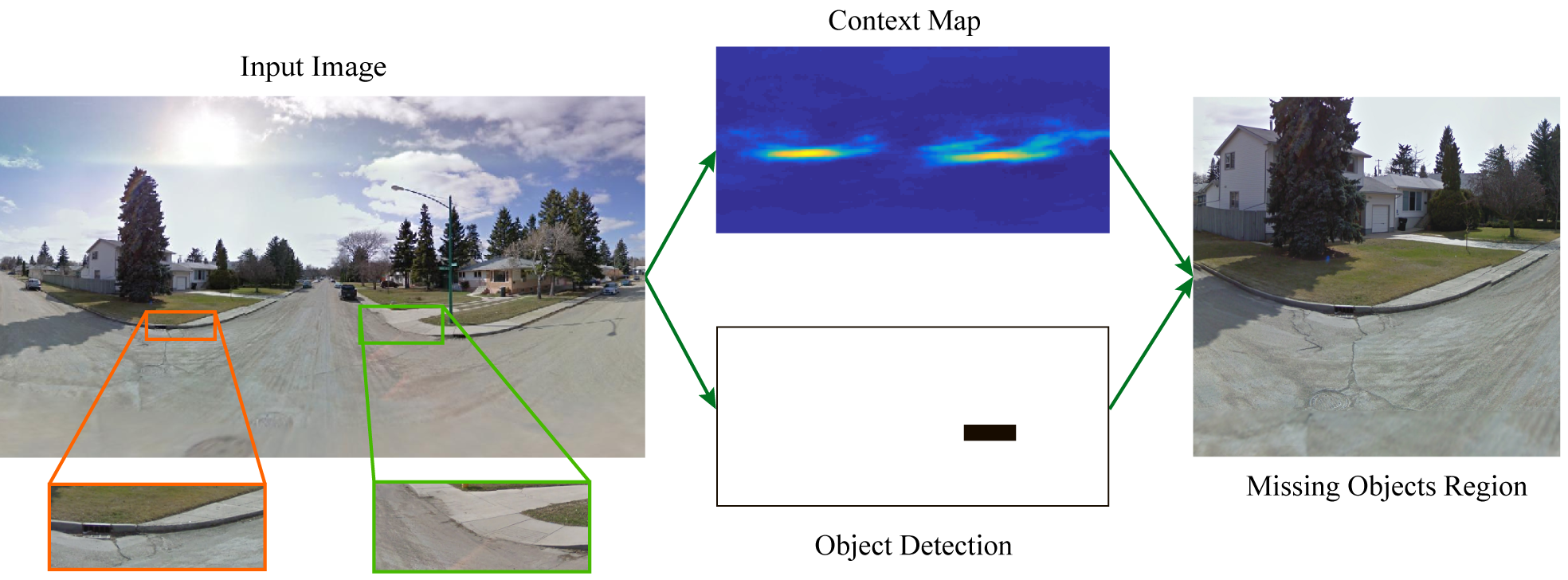}
    \captionof{figure}{When curb ramps (green rectangle) are missing from a segment of sidewalks in an intersection (orange rectangle), people with mobility impairments are unable to cross the street. We propose an approach to determine where objects are missing by learning a context model so that it can be combined with object detection results.}
    \label{fig:intro}
\end{center}%
}]


\begin{abstract}
Most of computer vision focuses on what is in an image. We propose to train a standalone object-centric context representation to perform the opposite task: seeing what is not there. Given an image, our context model can predict where objects should exist, even when no object instances are present. Combined with object detection results, we can perform a novel vision task: finding where objects are missing in an image. Our model is based on a convolutional neural network structure. With a specially designed training strategy, the model learns to ignore objects and focus on context only. It is fully convolutional thus highly efficient. Experiments show the effectiveness of the proposed approach in one important accessibility task: finding city street regions where curb ramps are missing, which could help millions of people with mobility disabilities.
\end{abstract}

\vspace{-2em}

\vfill
\begin{section}{Introduction}

Most fundamental computer vision tasks, e.g., image classification and object detection, focus on seeing what is there: for example, is there a curb ramp in this image, if yes, where is it? With the help of deep neural network models, computational approaches to such tasks are catching up to human performance in more and more benchmarks. However, humans can easily outperform algorithms in the task of inferring objects that are `not there': for example, is there a curb ramp in this image, if no, where \emph{could} it be?

We are interested in finding where objects are \emph{missing} in an image: an object of interest is not there, even though the environment suggests it should be. From a computational perspective, an object can be defined as missing in an image region when: 1) an object detector finds nothing; 2) a predictor of the object's typical environment, i.e. context, indicates high probability of its existence. Given an image, we want to detect all such regions efficiently. We summarize the relationship between an object detector and its context model in Table~\ref{tb:relation}. While there are many existing works on utilizing context in object detection (Section~\ref{sec:related-work}), they mainly focus on improving performance on finding typical objects with contextual and object information entangled. In this work we propose to train a standalone object-centric context representation to find missing objects. By looking at the reverse conditions, the exact same method can be adapted to find out of context objects too.

One practical motivation for finding missing objects comes from the street view curb ramp detection problem (Figure~\ref{fig:intro}). The task is to label curb ramps in a city's intersections so that people with mobility impairments can plan their route with confidence. 
Although existing work~\cite{hara_tohme:_2014} shows good performance in detecting constructed curb ramps, it cannot detect missing curb ramps.
Knowing this information is highly valuable: people with disabilities can assess the accessibility of an area; navigation algorithms can calculate better routes for pedestrians; the government can plan for future renovations accordingly. 
This is a very expensive and time consuming task for human labelers, which is partially the reason why such information is missing from public databases. Therefore, we are interested in developing an automatic algorithm that is effective and efficient. It can be used to scan a whole city for finding regions where curb ramps are missing. In this scenario, the number of found true missing curb ramp regions (recall) is more important than precision because it is much more light-weight to ask humans to verify algorithm results than to label images from scratch. Moreover, even if the algorithm reports one true missing curb ramp region but mistakenly ignores three others in an image, it is still valuable as a preprocessing step. With the missing curb ramp regions data, government can prioritize intersections in a city to send physical auditors in a more efficient way.

We believe the key to tackle this problem is to learn a model that focuses on context only and works efficiently just like an object detector: it scans each image and generates a probability heat map in which each pixel represents the probability that an object exists, even when no object is in sight. One big advantage of context and object decomposition is that we don't need abnormal object labels (missing/out-of-context) for training. A standalone context model can be learned from typical objects and later used for finding abnormal objects. This greatly simplifies training: normal objects are abundant and much easier to collect and label than abnormal objects.

\begin{table}
\centering
\begin{tabular}{c|c|c}
Object Score & Context Score & Image Region Remark\\
\hline
High   &   High    &  Typical objects\\ 
Low    &   High    &  Missing objects\\ 
High    &  Low     &  Out of context objects\\
\hline
\end{tabular}
\caption{Relationship between object and context. Object score is obtained from an object detector, while context score is from its context model.}
\label{tb:relation}
\end{table}

In this paper, we propose such a model based on convolutional neural networks and a novel training strategy to learn a standalone context representation of a target object. 
We start by introducing a base network in Section~\ref{sec:learn-mask}. It takes input images with explicit object masks and learns useful context from the remaining areas of the images. Because of the limitations discussed in Section~\ref{sec:learn-and-ignore}, we then propose a fully convolutional version of the network that learns an implicit object mask such that it ignores objects in an image and focuses purely on context. It does not require object masks during test time. Finally, Section~\ref{sec:pipeline} describes the procedure of using the context model to find missing objects regions.

The contributions of this work are as follows. First, we propose a method to learn an object-centric context representation by learning from object instances with masks. Second, we propose a training strategy to force the network to ignore objects and learn an implicit mask. The model is fully convolutional so it also speeds up probability heat map generation significantly. Finally we present promising results on missing curb ramps detection problem in street view images, and a preliminary result on finding out-of-context faces.

\end{section}

\begin{section}{Related Work}\label{sec:related-work}

\noindent\textbf{Context in Object Recognition}. Cognitive science studies have shown a large body of evidence that contextual information affects human visual search and recognition of objects~\cite{OlivaT07tcs,bar_visual_2004}. In computer vision, recently it also has become a well accepted idea that context helps in object recognition algorithms~\cite{divvala_empirical_2009,mottaghi_role_????,ouyang_single-pedestrian_2015,wang_whats_2016}. Usually, context is represented as the semantic labels around an object. \cite{rabinovich_objects_2007} uses a Conditional Random Field to model contextual relations between objects' semantic labels to post-process object recognition results. \cite{mottaghi_role_????} builds a deformable part model that incorporates context labels around an object as `parts'. Because of the coupling between context and object information, these methods are unsuitable to detect missing object regions.

Torralba et al. proposed the Context Challenge~\cite{torralba_contextual_2003} that consists in detecting an object using exclusively contextual information. They take the approach of learning the relation between global scene statistical features and object scale and position. Visual Memex~\cite{malisiewicz_beyond_2009} is a model that can either retrieve exemplar object instances or predict semantic identity of a hidden region in an image. It uses hand-crafted features and models context as inter-category relations. Our approach can be seen as a general approach that attempts to address this challenge, without the need for designing hand-crafted features or preset object classes.

\noindent\textbf{Finding Missing Objects}. Grabner et al. proposed to use the General Hough Transform to find objects that are missing in some frames during object tracking~\cite{grabnerMGG10cvpr}. The idea is to estimate the position of a target object from surrounding objects with coupled motions.

\noindent\textbf{Computer Vision with Masked Images}. Recently Pathak et al.~\cite{pathakCVPR16context} proposed to learn a convolutional neural network context encoder for image inpainting. Both their work and ours train convolutional neural networks with masked images. But the purpose is very different as they try to learn a generative model to inpaint the mask while we learn a discriminative model to infer what is inside the mask. Also, our work is capable of using a much more efficient fully convolutional structure.

\noindent\textbf{Accessibility Task}. With massive online resources such as the Google Street View service, many computer algorithms are designed to help people with disabilities and improve their quality of life. CrossingGuard~\cite{guy_crossingguard:_2012} is a system designed to help visually impaired pedestrians to navigate across intersections with help from Amazon Mechanical Turk. Tohme~\cite{hara_tohme:_2014} is a semi-automated system that combines crowdsourcing and computer vision to collect existing curb ramp positions in city intersections using GSV images. It uses the Deformable Part Models~\cite{felzenszwalb_object_2010} as a curb ramp detector and asks Mechanical Turkers to verify the results. They provide a street view curb ramp dataset with 1086 city intersection images, which we use in our experiment.

\end{section}

\begin{figure*}[!ht]
    \begin{center}
\includegraphics[width=\textwidth]{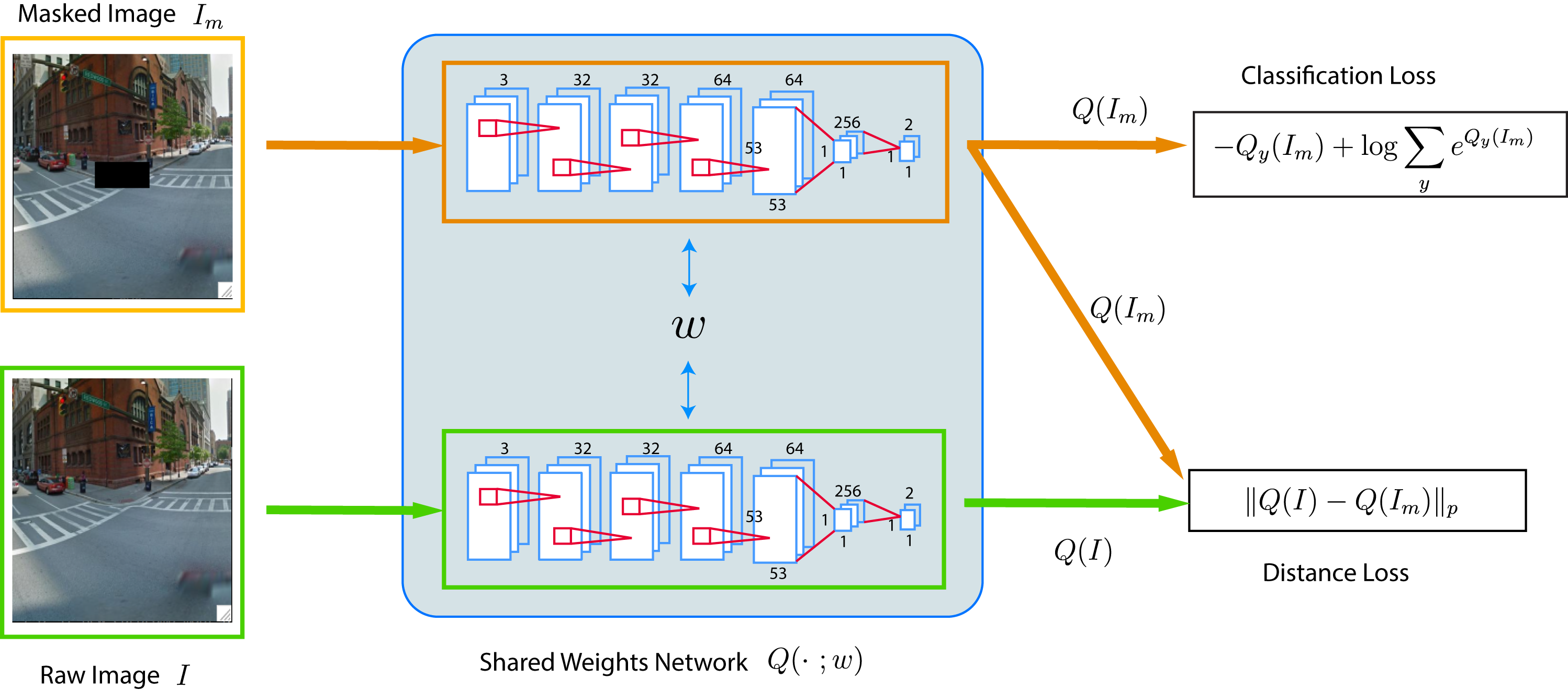}
    \end{center}
    \caption{Training scheme of the Siamese trained Fully convolutional Context network (SFC). The intuition is to enforce the fully convolutional network $Q$ to output similar result regardless of whether an image is masked or not. Additionally, the network should produce correct classification label. The training is done in a Siamese network setting with shared weights $w$.}
    \label{fig:context-network}
\end{figure*}

\begin{section}{Learning Context from Explicit Object Masks}\label{sec:learn-mask}
    In this section, we introduce the base version of the proposed context learning algorithm.
If `context' is considered to be anything that surrounds an object except for the object itself, this model is learning context literally: any target object instances in training images are masked out. Here we assume an object's visual extent is fully represented by its bounding box.

We train this context model in a binary image classification setting. Positive samples are collected so that each image has an object at its center, with a black mask (value equals zero after preprocessing) covering the object's full extent. The ratio between an object's bounding box width and the whole image width is about 4.0 with the purpose of including a large contextual area. Negative samples are random crops with a similar black mask at center. The position of the negative crops is chosen so that the masked region will not cover any groundtruth labeled objects with more than a Jaccard index~\footnote{Defined as the intersection-over-union ratio of two rectangles.} of 0.2.

When there are multiple object instances in the image, we only mask out one object at a time for positive samples. This is because the existence of other object instances could be useful context: for example, curb ramps often appear in pairs.

To prevent the model trivially learning the particular mask dimension, we force the negative samples to have a similar distribution of mask dimensions as the positive samples. The sampling strategy is to interleave the positive samples and negative samples, and use the previous positive sample's mask dimension in the next negative sample.

We train a convolutional neural network model $Q$. The network consists of four convolutional layers with pooling and dropout, and two fully connected layers. Its structure is summarized in Table~\ref{tb:network-structure}. Cross entropy loss (Eq.~\ref{eq:Lc}) is used as the classification loss:

\begin{equation}
\mathcal{L}_c = -Q_y(I_m) + \log \sum_y e^{Q_y(I_m)},
\label{eq:Lc}
\end{equation}
where $y \in \{1,2\}$ is the groundtruth label for a masked image $I_m$ (1 for positive, 2 for negative), $Q(I_m)$ is a 2x1 vector representing the output from the network $Q$, while $Q_y(I_m)$ represents its $y$-th element.

\begin{table}[!ht]
    \centering
\begin{tabular}{c|c|c}
    \hline
    Layer (type)           &          Shape     &     Param \#\\
    \hline
    Convolution2D & (3, 3, 32) & 896 \\
    Convolution2D & (3, 3, 32) & 9248 \\
    MaxPooling2D   & (2, 2) & 0    \\
    Dropout            &  - & 0     \\
    Convolution2D & (3, 3, 64) & 18496       \\
    Convolution2D & (3, 3, 64) & 36928     \\
    MaxPooling2D   & (2, 2) & 0         \\
    Dropout             & - & 0         \\
    FullyConnected & (53*53*64, 256) & 46022912 \\
    Dropout             & - & 0        \\
    FullyConnected & (256, 2)& 514       \\
    \hline
    \multicolumn{3}{c}{Total params: 46,088,994}\\
    \hline
\end{tabular}
\vspace{3px}
\caption{Neural network structure summary for the base network. Convolution filter shape is represented by (filter width, filter height, number of filters) tuple. The network expects to take an input image of size 224x224, with explicit mask at the center.}
\label{tb:network-structure}

\vspace{3px}
\begin{tabular}{c|c|c}
    \hline
    Convolution2D & (53, 53, 256) & 46022912 \\
    Dropout             & - & 0        \\
    Convolution2D & (1, 1, 2)& 514       \\
    \hline
\end{tabular}
\vspace{3px}
\caption{Fully convolutional layers to substitute for the last three layers of the base network. This network can take arbitrary sized input, with no explicit mask needed.}
\label{tb:network-structure-fc}
\end{table}

During test time, a sliding window approach is used to generate the probability heat map for a new image so that each pixel has a context score of how likely it is to contain an object. At each position, a fixed size (224x224 in our implementation) image patch is cropped with the center region masked out to be fed into the base network. The size of the center mask region is chosen based on the statistics of object bounding boxes from the training set. 

\end{section}

\begin{section}{A Fully Convolutional Model that Learns Implicit Masks}\label{sec:learn-and-ignore}

There are several issues with a network trained with masked images. First, the network tends to learn artifacts. For example,~\cite{pathakCVPR16context} reports that training with rectangular mask makes the network learn ``low level image features that latch onto the boundary of the mask''. They propose to use random mask shapes to prevent this issue. However, we cannot use the same strategy for this task because our mask is strictly tied into the visual extent of an object.
Second, during testing time, the network is expecting to see every input with an explicit mask. The efficiency of this operation becomes an issue when we have to evaluate the network at all possible positions and scales to generate a heat map. There are standard procedures to convert a convolutional neural network with fully connected layers into a fully convolutional one~\cite{sermanet2013overfeat}, so the evaluation is much efficient for images of arbitrary size. However, in our case the situation is complicated. During training, the network always sees input images with all zeros at the center, so the weights of neurons with receptive field on this region can be arbitrary because no gradients are updated. If we apply the converted fully convolutional network to unmasked images, outputs from those neurons can affect the network's output arbitrarily.

The question is then, can we train a network so that it is fully convolutional and learns context by ignoring the masked region `by heart'?

The answer is yes and we now propose a training strategy to make a network learn the implicit mask. The intuition is that we want the network to output similar results regardless of whether the image is masked or not. By enforcing this objective, the network should learn to find visual features that are shared in both masked and raw images: i.e. from the unmasked regions.

Formally, we want to minimize a distance loss in addition to the classification loss used in the base network:
\begin{equation}
\mathcal{L}_d = || Q(I_m) - Q(I) ||_p,
\end{equation}
where $Q(I_m)$ is the output vector from the network $Q$ with masked image $I_m$ as input, $Q(I)$ is the output vector from $Q$ with the unmasked raw image $I$ as input, and $\| \cdot \|_p$ represents the $L_p$-norm. 

Effectively, we have two shared-weight networks that are fed with masked and raw image pairs (Figure~\ref{fig:context-network}). The network is a fully convolutional version of the base network (Table~\ref{tb:network-structure-fc}). One stream of the network computation takes the masked image as input and outputs $Q(I_m)$. In parallel, the other stream of network computation takes the unmasked raw image as input and outputs $Q(I)$. The classification loss $\mathcal{L}_c$ is calculated based on $Q(I_m)$ alone, while the distance loss $\mathcal{L}_d$ is calculated by $Q(I_m)$ and $Q(I)$. This structure is known as a Siamese Network~\cite{chopra_learning_2005} so we call this network as the Siamese trained Fully convolutional Context (SFC) network. Following~\cite{chopra_learning_2005}, we choose the $L_1$ norm in distance loss $\mathcal{L}_d$. We expect the SFC network to learn an implicit object mask by assigning zero weights to neurons whose receptive field falls onto the center object mask region of an input image. During test time, unlike the base network, we don't have to manually set the mask size: the SFC network has encoded this information in convolutional filters' weights.

Finally, the overall training objective is defined as a weighted sum of the two losses:

\begin{equation}
\mathcal{L} = \lambda \mathcal{L}_d + \mathcal{L}_c,
\end{equation}
where $\lambda=0.5$ in our implementation.

The benefits of this training strategy are three fold:

1) Because the SFC learns to ignore the object mask region, we can directly apply it to new unmasked images with arbitrary size, so it is highly efficient to generate a dense probability map. Figure~\ref{fig:efficient} shows a comparison between heat maps generated by the base network and the SFC network. An image with size 1024x2048 takes about 5 minutes to generate a heat map with the base network while the SFC network takes less than 4 seconds to generate a map with higher spatial resolution.

2) The SFC network is less prone to artifacts. It is possible for the base network to learn artifact features along the boundary of masks. Because such features are not present in unmasked images, the SFC network ignores them to reduce training distance loss $\mathcal{L}_d$.

3) During training, we can perform hard negative mining efficiently. Between each training epoch, we can apply the SFC network on all training images to generate heat maps and find high score false positive regions. Because of the efficiency of fully convolutional networks, this step can be easily included in training. Section~\ref{sec:exp/curb} shows that hard negative mining indeed improves the network performance by a large margin.

\begin{figure}[!ht]
    \centering
\includegraphics[width=3in]{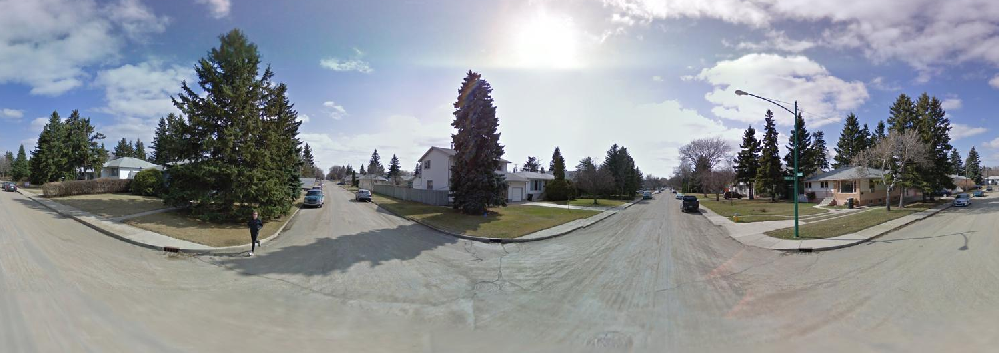}
\includegraphics[width=3in]{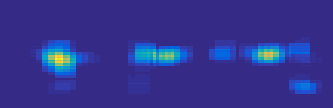}
\includegraphics[width=3in]{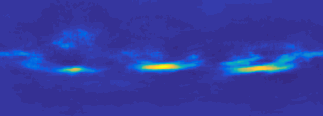}
\caption{Top: an input street view panorama image. Middle: the heat map generated by the base network using a sliding window approach. It has low spatial resolution due to high cost of the computation. Bottom: dense heat map generated by the SFC network.}
\label{fig:efficient}
\end{figure}

\end{section}

\begin{section}{Finding Missing Object Regions Pipeline}
\label{sec:pipeline}

With a trained standalone context network (base network or SFC network), we summarize the procedure for finding missing object regions in the following steps. 

1) Generate a context heat map using the context network $Q$ for a test image. The context heat map shows where an object should occur in the image.

2) Generate object detection results using an object detector. 
Convert object detection bounding boxes into a binary map by assigning 0 to the detected box region, 1 otherwise. This binary map shows where objects have already been found in the image. We want to find the regions where no objects are found.

3) Take an element-wise AND operation between context heatmap and the binary map. The resulting map shows the places in which an object should occur according to context but in which the detector has found none.

4) Retrieve the high scored regions (above a preset threshold) according to the resulting map, crop them from the original image. These are the regions where objects are missing.
\end{section}

\begin{section}{Experiments}

In this section, we first examine the characteristics of the base network and the SFC network in Subsection~\ref{sec:exp/understand}. Then we evaluate their effectiveness. With the decomposition of context and object information, we study two unique tasks that can be efficiently performed with a standalone context model. Subsection~\ref{sec:exp/curb} shows experimental results of detecting missing curb ramp regions in street view images. Subsection~\ref{sec:exp/face} shows preliminary results of detecting faces that are out of context in unconstrained images. 

\begin{subsection}{Characteristics of the Trained Model}\label{sec:exp/understand}

As a validation study, we first check the sensitivity of the base and SFC networks with regard to small changes in the input image. All experiments are conducted on the curb ramp street view dataset. The desirable model should have small response variation to the center region of the input image, where the mask was put during training. For each image, we change one pixel value at a time, by adding a small noise. The $L_2$ distance between a network's output before and after the disturbance is recorded for each pixel. In the end we obtain a map that shows which region in the image has large impact on the network's output. This can be seen as an estimate of the first order derivative of the network with respect to its input. Figure~\ref{fig:sensitivity} shows the result of this experiment with comparison between the base network and SFC network. The result is summed over 20 different image samples.

\begin{figure}[!ht]
    \centering
\includegraphics[width=1.5in]{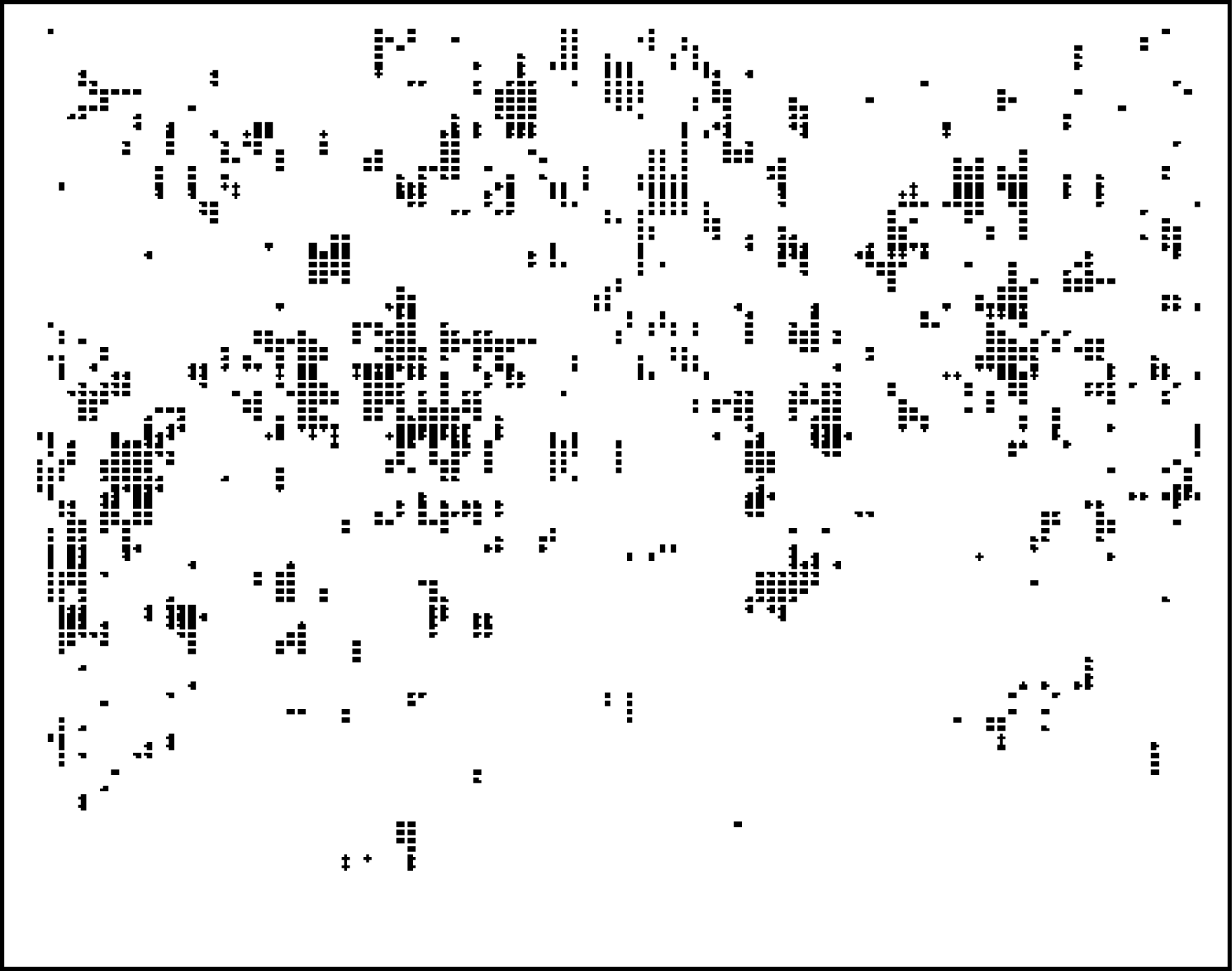}\hspace{5px}
\includegraphics[width=1.5in]{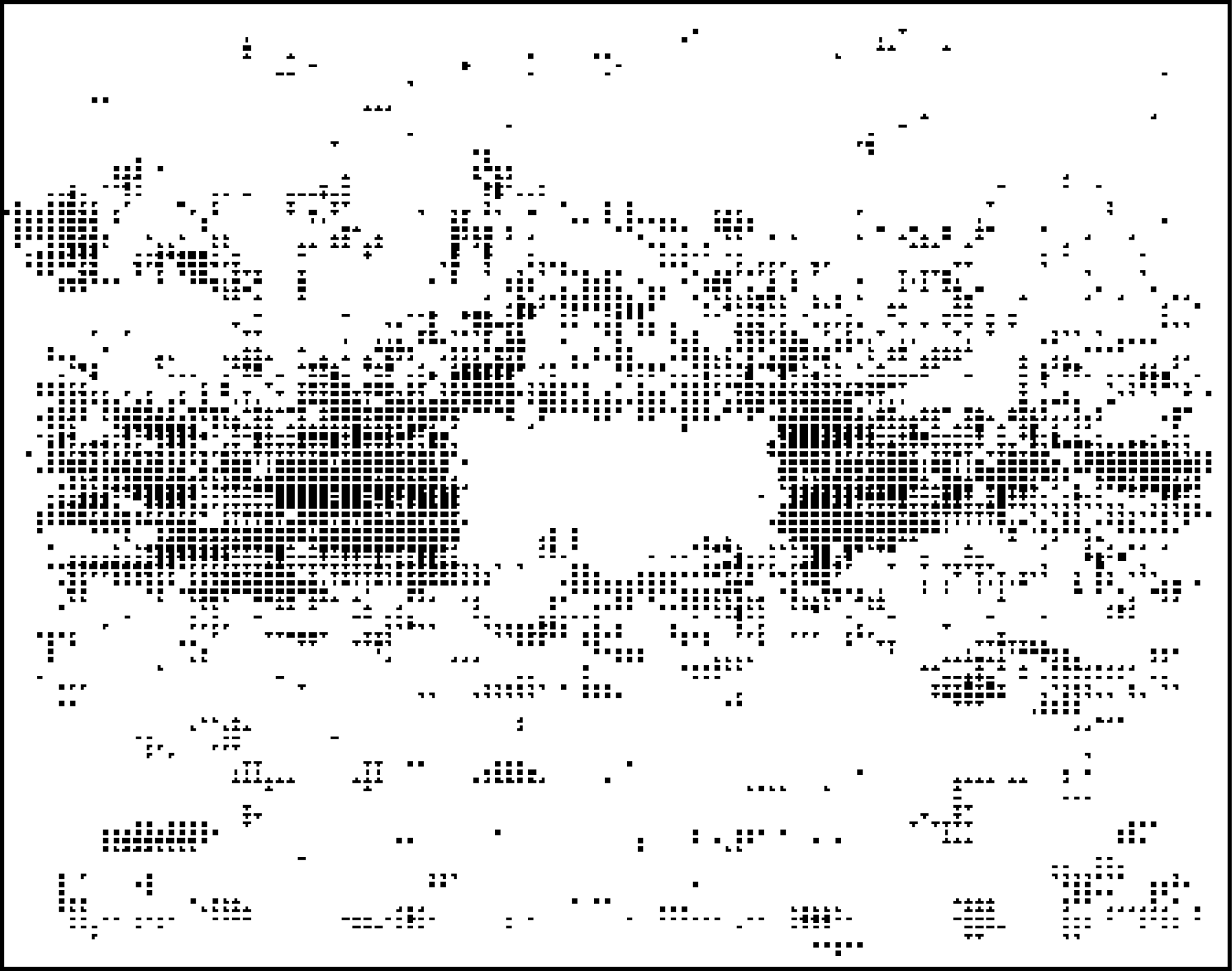}
\caption{The sensitivity map of the base network (left) and the SFC network (right): a dark dot indicates high sensitivity spot. Compared to the base network, the SFC map has a clear blank area at the center, which indicates that changes in this region have little effect on the network's output. The SFC network learns an implicit region mask.}
\label{fig:sensitivity}
\end{figure}

From the result it is clear that the SFC network has small sensitivity at the center region of the input image. This is most likely due to the network learning to mute neurons whose receptive field falls at the center region of the input image. On the other hand, the base network shows no such preference. The blank region in the SFC's sensitivity map can be seen as a visualization of an approximation to the learned implicit region mask. 

Next we check the distance loss $\mathcal{L}_d$ of the base network and the SFC network on test data. Following the same set of training hyper-parameters and setup (learning rate, training epochs) to train the two networks, the mean $\mathcal{L}_d$ loss is summarized in Table~\ref{tb:l1-diff}. It is clear that the SFC network is much more consistent in producing similar outputs regardless of the object masks.

\begin{table}[!ht]
\centering
\begin{tabular}{c|c|c}
\hline
~ & SFC network & Base network\\
\hline
$\mathcal{L}_d$ loss &   0.041 & 2.27\\
\hline
\end{tabular}
\vspace{3px}
\caption{Mean $\mathcal{L}_d$ loss of the two networks on the curb ramp dataset test set. Lower loss means smaller differences between the network's output from masked and unmasked images.}
\label{tb:l1-diff}
\end{table}

The above experiments have demonstrated that the SFC network works just as we have expected: 1) it learns an implicit mask so it is less sensitive to any changes in the center region; 2) the useful features that it learns for the classification task are mainly from the unmasked regions.

\end{subsection}

\begin{subsection}{Finding Missing Curb Ramp Regions}\label{sec:exp/curb}

\noindent\textbf{Setup.}  We want to find missing curb ramps in the street view curb ramps dataset~\cite{hara_tohme:_2014}. The dataset contains 1086 Google Street View panoramas which come from four cities in North America: Washington DC, Baltimore, Los Angeles and Saskatoon (Canada). Each panorama image has 1024x2048 pixels. It provides bounding box labels for existing curb ramps. On average there are four curb ramps per image. The dataset also contains bounding box labels for missing curb ramps regions.

The dataset is split into half training and half testing. Each image is converted to YUV color space and normalized to be zero mean and one standard deviation in all channels.
We use the curb ramp detector provided with the dataset, a Deformable Part Model, with default settings.

\noindent\textbf{Training.} For each epoch, 5000 samples are generated from training data, with half positives and half negatives. Figure~\ref{fig:training-pairs} shows several examples. Each sample has 50\% probability of being horizontally flipped for data augmentation purposes. Positive samples contain useful contextual information around the curb ramps. Negative samples are sampled randomly from the remaining areas of the panoramas. To train the SFC network, each sample is prepared with two versions: raw and masked. We resize positive samples such that the object width is close to 55 pixels in a 224 pixels wide image. Each negative sample uses the same object mask and scale as the last positive sample to prevent the network overfitting to the mask dimension distribution.

\begin{figure}[!ht]
    \centering
\includegraphics[width=3in]{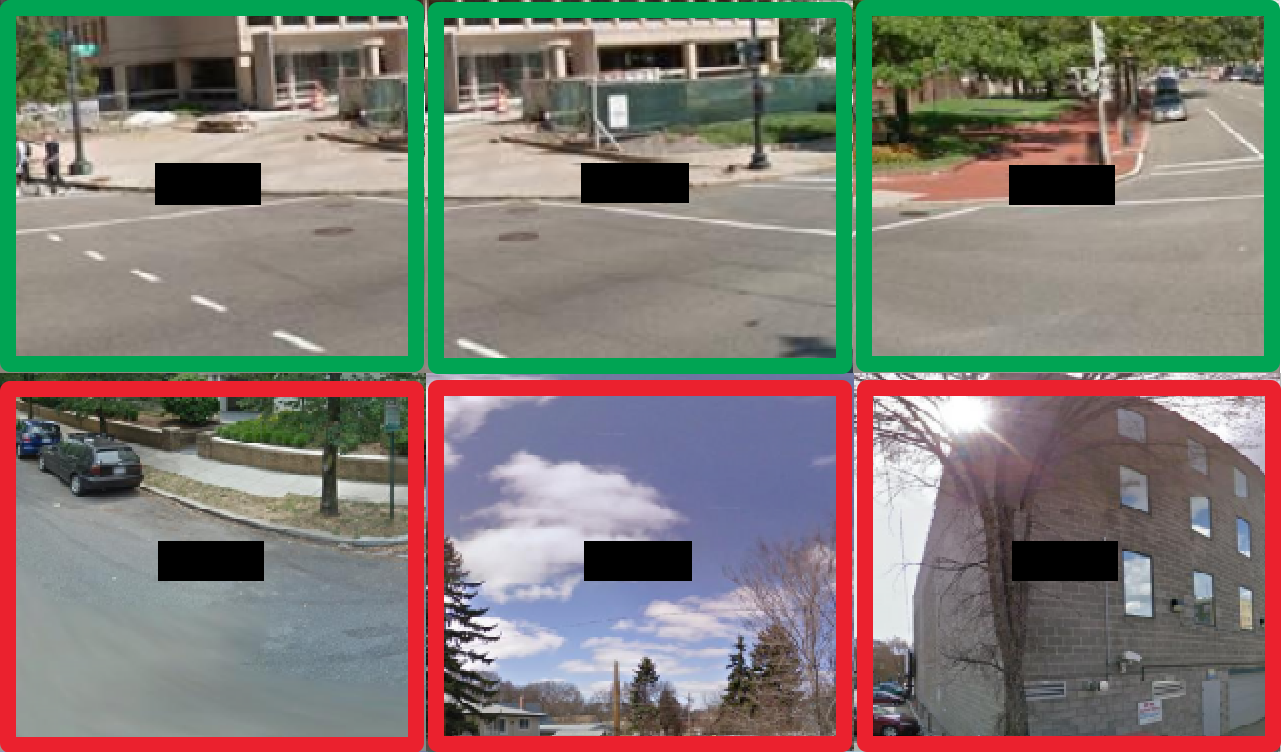}
\caption{Training examples of curb ramps. Green rectangles represent positive samples, red rectangles represent negative samples.}
\label{fig:training-pairs}
\end{figure}

We use the Keras/Tensorflow software package~\cite{keras} to train the network models. The optimization algorithm uses Adadelta with default parameters. Since this is an adaptive learning rate method, there is no need to set a learning rate schedule during training. 20\% of the training data is used as a validation set for an early stopping test.

We have trained a base network and a SFC network using the same hyper-parameters and training setup.

\noindent\textbf{Results.} Following the procedure described in Section~\ref{sec:pipeline}, we run the two networks on test images to generate probability heat maps of where curb ramps should be in the image. Each heat map for the base network is generated in a sliding window scheme with a stride of 10 pixels, and various object mask widths of \{50, 70, 100\} to generate multi-scale maps. The SFC network doesn't need an object mask size, so we resize the input panorama image with scales \{0.5, 0.7, 1.0\}. The numbers are chosen so that two networks see similar image pyramids. We use the DPM detector provided with the dataset to generate detection results. For each panorama, we generate a final map that combines detection and context map and retrieve the high scored regions (above certain threshold) with size $d\times d$ from the raw image. According to preliminary empirical studies, we set the context threshold to 0.4 throughout the experiment.

We use human verification to evaluate the quality of the reported missing curb ramp regions. For that purpose, we develop a web based interface (Figure~\ref{fig:exp/webpage}) that displays a gallery of found regions, ranked by their context scores. For each candidate region, the user provides feedback on whether it is truly a region with missing curb ramps. We compare context maps generated by the base and the SFC networks with three baseline methods: random scores, spatial prior map, and a Faster RCNN~\cite{ren2015faster} based missing curb ramp detector.

\begin{figure}[!ht]
    \centering
    \includegraphics[width=3.2in]{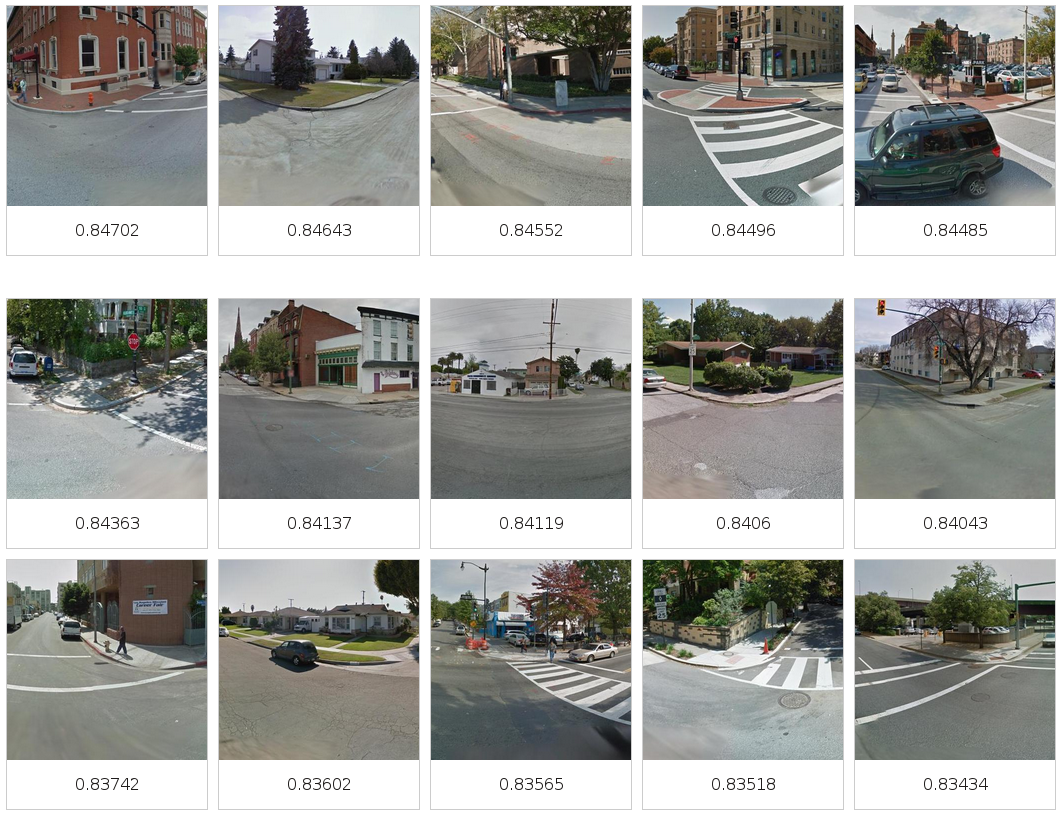}
    \caption{The web interface for verification. Each thumbnail shows one retrieved region, with its score displayed below. A user clicks on a thumbnail to label it as a true missing curb ramp region.}
    \label{fig:exp/webpage}
\end{figure}

Random scores assigns uniformly random context scores from $[0,1]$ to all positions in an image. This is a reference baseline showing the performance by chance.

A spatial prior map is built using the prior positions of curb ramps in street view panoramas. We use the prior map as a replacement for the context map for comparison. We collect the prior spatial distribution of all curb ramps from the training images. The collected distribution is smoothed with a 30x30 pixel Gaussian kernel with sigma=10. Figure~\ref{fig:exp/priormap} shows the spatial prior map used in our experiment. Because most panoramas are at street intersections, there is strong spatial structure consistency among the dataset. We expect this approach to be a reasonable baseline.

\begin{figure}[!ht]
\centering
    \includegraphics[height=1in,width=3in]{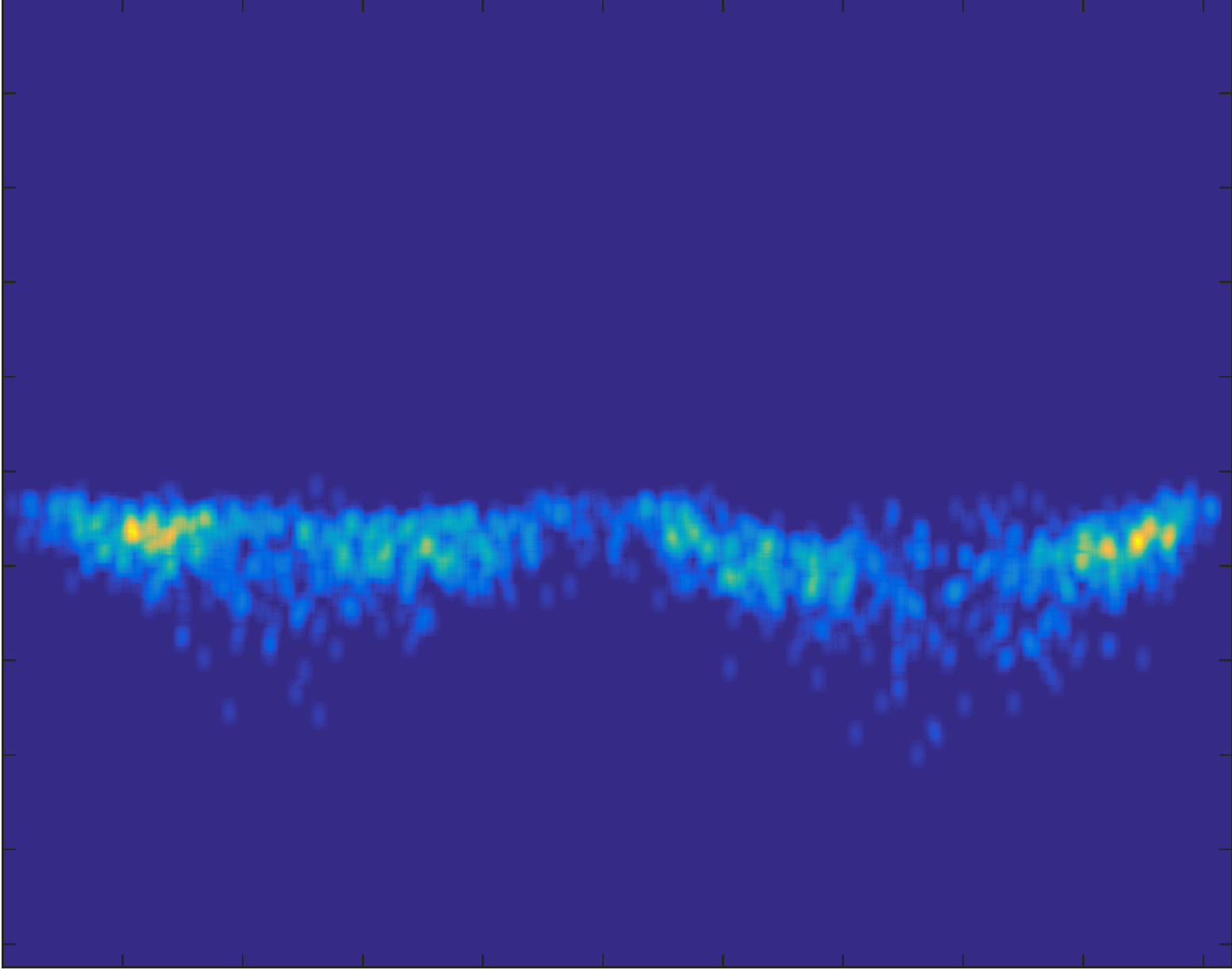}
    \caption{Spatial prior heat map generated from the groundtruth locations of curb ramps in the training set. There are strong spatial structure similarity shared among the dataset.}
    \label{fig:exp/priormap}
\end{figure}

With missing curb ramp regions labels, we can treat this task as a standard object detection problem and directly train a Faster RCNN detector: the positive `object' is a region labeled as missing curb ramps. Note that a Faster RCNN detector is capable of learning context because it's an end-to-end approach: potentially the detector can learn from the whole image to predict locations of missing curb ramp regions. We expect the Faster RCNN detector to be a strong baseline.

The verification of the missing curb ramp regions requires domain knowledge. We asked one researcher who has extensive experience with accessibility problems to verify the results using our web interface. Figure~\ref{fig:exp/plots} shows the comparison in recall of true missing curb ramp regions versus number of visited regions (Recall@K). The retrieved region size is set to $d=400$ pixels. 500 regions were retrieved from 543 test images. 

The result shows that the SFC network with hard negative mining outperforms all other methods. We believe its superiority comes from the highly efficient fully convolutional structure that helps in training and generating high resolution context maps. Spatial prior map shows reasonable performance, which confirms the spatial bias of curb ramps locations in the dataset. Unlike the spatial prior map, the proposed methods can work well on other datasets that have no such bias. The Faster RCNN detector has significantly less recall compared with the SFC networks. With more missing curb ramp regions as training data, we expect the Faster RCNN detector to show improved performance; on the other hand, the SFC network does not even need missing curb ramp labels in training. The proposed methods learns useful context information from normal curb ramps, which are much easier to collect and label than missing curb ramp regions. Moreover, the SFC network is using detection results from a less advanced curb ramp detector (a DPM model shipped with the dataset): 77\% of the false missing curb ramp retrievals are due to inaccurate curb ramp detections. Due to the page limit, we show more qualitative results of retrieved regions by these methods in the supplementary document.

\begin{figure}[!ht]
\centering
\includegraphics[width=3in]{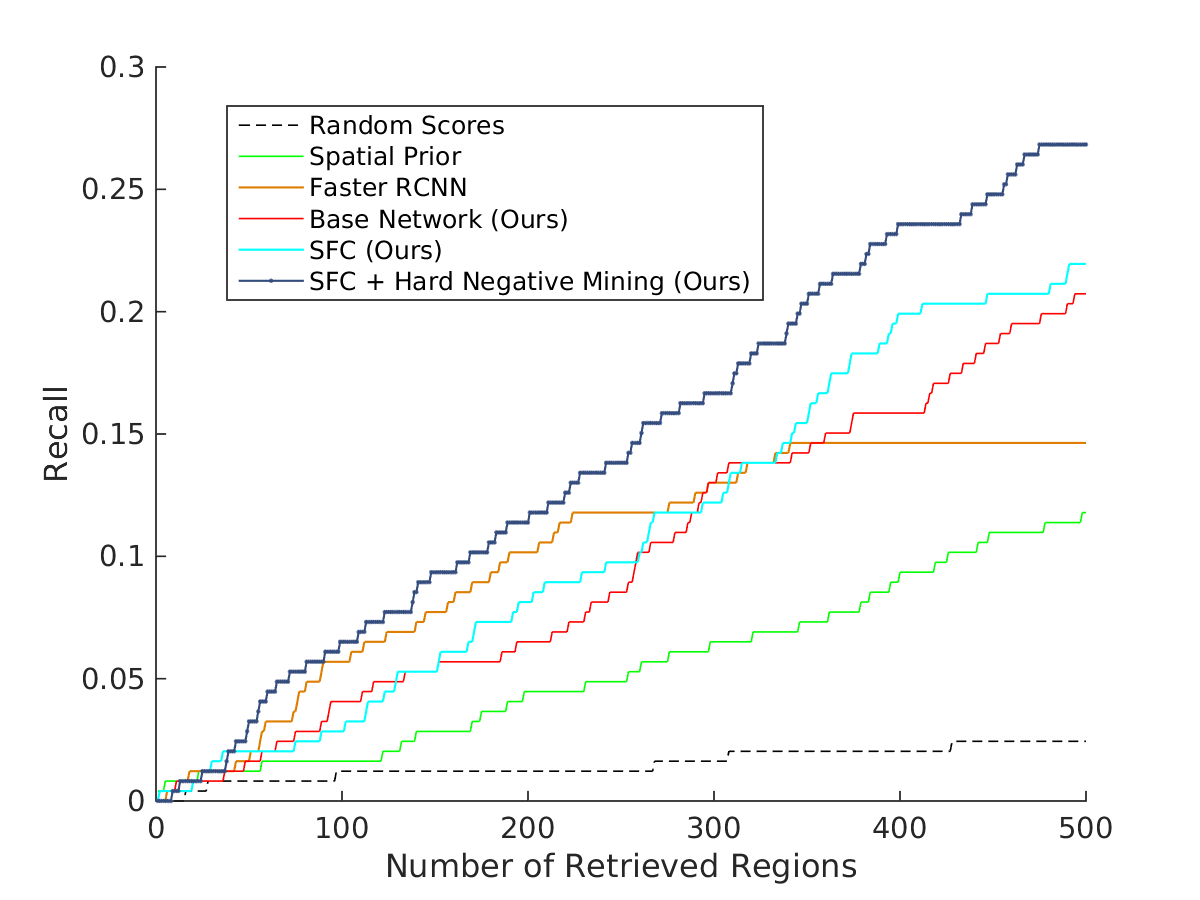}
\caption{Recall of true missing curb ramp regions vs number of regions viewed (Recall@K). Our base and SFC networks outperform the two baseline methods (random scores and prior maps) by a large margin. The difference of recall between the Faster RCNN detector and the proposed method is significant. The SFC network with hard negative mining has the best result among the proposed methods.}
\label{fig:exp/plots}
\end{figure}

Additionally, we investigate the effects of the retrieved region size $d$ on number of true missing curb ramp regions. Specifically, we vary the cropped region size from 400 pixels in width to 100 pixels. With smaller region size, it becomes crucial that the region is accurately localized with missing curb ramps at the center. Table~\ref{tb:vary-size} shows that the SFC network is not affected too much by the reduced field of view. This is because the regions it found are very well localized (See Figure~\ref{fig:exp/webpage}). On the other hand, two baseline methods (random scores and prior maps) are performing poorly when the region size becomes small. Note that smaller windows can lead to ambiguities, which can result in `falsely admitted' missing curb ramps due to human verification error. This is reflected in Table~\ref{tb:vary-size} first row: from region width 400 to 200, the SFC performance goes up.

\begin{table}
\centering
\begin{tabular}{c|ccc}
\hline
Region Width      & 400 & 200 & 100 \\
\hline
SFC   & 21.8  & 24.1  & 21.6 \\  
Spatial Prior & 10.2  & 6.8  & 3.2  \\ 
Random Scores & 3.8   & 1.6  & 0\\ 
\hline
\end{tabular}
\caption{Effect of retrieved region size on the mean number of found missing curb ramps with 255 regions (higher the better). As the region width shrinks, SFC performs very consistently while the two baseline methods (random scores and prior maps) suffer from poor localization. }
\label{tb:vary-size}
\end{table}

\noindent\textbf{Discussion.} Among the 543 street view intersections in the test set, we are able to find as many as 27\% missing curb ramp regions using the proposed method by merely looking at 500 regions. This is an impressive result for the following reasons. 1) The whole process is very efficient (Table~\ref{tb:timecost}) such that it can be easily deployed to scan new city areas. For example, there are about 2,820 intersections in Manhattan, New York: it will take just a few hours for our system to find missing curb ramps in a region with 1.6 million population. 2) Research has shown that curb ramps condition (missing or not) shows high proximity consistency: if one intersection has missing curb ramps, it is highly likely that the intersection nearby has similar issue. Results from our system can be used as an initial probe to quickly find city areas that need special attention. 

\begin{table}
\begin{tabular}{c|ccc}
\hline
~ & Context Map (*)   & Detection & Verification \\
\hline
Cost &  4s/image    &    22s/image  &   20min/500 ims\\
\hline
\end{tabular}
\caption{Time cost for different steps in finding missing curb ramps. The whole process is efficient as context map and detection can be generated in parallel. *Using the SFC network.}
\label{tb:timecost}
\end{table}

\end{subsection}

\begin{subsection}{Finding Out of Context Faces}\label{sec:exp/face}

The pipeline in Section~\ref{sec:pipeline} for finding missing objects can be adapted to find out of context objects with just a few small modifications: change step 2 by assigning 1 to detected box regions and 0 for other regions; change step 4 to retrieve the lowest scored regions. Here we show a simple preliminary result of finding out of context faces to demonstrate both the generalization ability of the proposed method on different domains and the possible future directions.

The task is to find out of context faces in the Wider face dataset~\cite{yang2016wider}. Using a similar procedure as in finding missing objects and a state-of-the-art face detector~\cite{zhang_joint_2016}, we retrieve the top 500 face regions from the validation set that contain high face detector score and low context score. For evaluation purposes, we define an out of context face as a face without visible support from a body. Figure~\ref{fig:exp/face} shows qualitative results of found out of context faces using the SFC network. We compare the SFC network result with random scoring. Out of 500 regions, the SFC network can find 27 out of context faces while random scoring found 14. While the result is preliminary, it suggests that the proposed method has the potential to be used in many other applications where finding out of context objects is important: for example, visual anomaly detection.

\begin{figure}[!ht]
    \centering
\includegraphics[width=3in]{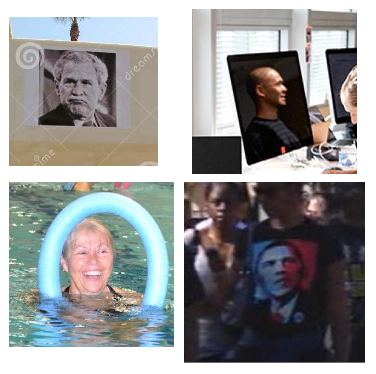}
\caption{Retrieved out-of-context faces with a SFC network from the Wider face dataset.}
\label{fig:exp/face}
\end{figure}

\end{subsection}

\end{section}
\begin{section}{Conclusion}

We present a approach to learn a standalone context representation to find missing objects in an image. Our model is based on a convolutional neural network structure and we propose ways to learn implicit masks so that the network ignores objects and focuses on context only. Experiments show that the proposed approach works effectively and efficiently on finding missing curb ramp regions. 
\end{section}

\section*{Acknowledgments}

This work was supported by an NSF   grant   (IIS-1302338).

{\small
\bibliographystyle{ieee}
\bibliography{library}
}

\end{document}